\documentclass[final]{cvpr}

\usepackage{times}
\usepackage{epsfig}
\usepackage{graphicx}
\usepackage{amsmath}
\usepackage{amssymb}
\usepackage{xcolor}
\usepackage{color}
\usepackage{tabularx} 
\usepackage{subfigure}
\usepackage{float} 

\definecolor{purple}{rgb}{0.75, 0.0, 1.0}

\newcommand{\ddi}{{D$^2$IM-Net}}
\newcommand{\ddg}{{D$^2$IM-Net$_{GL}$}}

\usepackage[pagebackref=true,breaklinks=true,colorlinks,bookmarks=false]{hyperref}

\begin{document}
\title{\ddi: Learning Detail Disentangled Implicit Fields from Single Images}

\author{
Manyi Li
\qquad
Hao Zhang\\ 
Simon Fraser University
}

\maketitle
\begin{abstract}
We present the first single-view 3D reconstruction network aimed at recovering geometric details from an input image which encompass both topological shape structures and surface features. Our key idea is to train the network to learn a {\em detail disentangled} reconstruction consisting of two functions, one implicit field representing the coarse 3D shape and the other capturing the details.
Given an input image, our network, coined \ddi{}, encodes it into global and local features which are respectively fed into two decoders. The base decoder uses the global features to reconstruct a coarse implicit field, while the detail decoder reconstructs, from the local features, two displacement maps, defined over the front and back sides of the captured object.
The final 3D reconstruction is a fusion between the base shape and the displacement maps, with three losses enforcing the recovery of coarse shape, overall structure, and surface details via a novel {\em Laplacian\/} term. 
\end{abstract}

\section{Introduction}
\label{sec:intro}

Reconstructing 3D shapes from single-view RGB images is the prototypical ill-posed problem in computer vision. Recently, rapid advances in deep learning have propelled the development of data-driven single-view 3D reconstruction methods. In particular, the emergence of {\em neural implicit models\/}~\cite{chen2019learning,park2019deepsdf,mescheder2019occupancy} for 3D shape representation learning has led to much improved reconstruction quality compared to methods designed for voxel grids, meshes, and point clouds. However, while technically the implicit fields could be sampled to an arbitrarily high spatial resolution, state-of-the-art reconstruction methods still are unable to adequately recover fine-level {\em geometric details}.

\begin{figure}
\centering
\includegraphics[width=\linewidth]{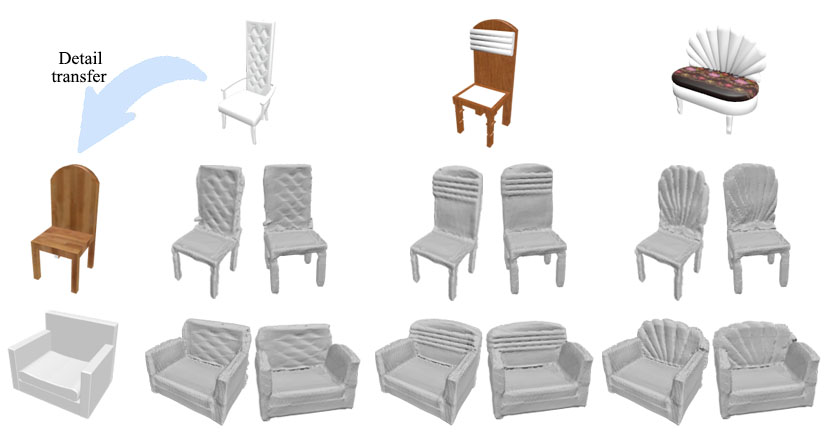}
\caption{Our network learns to reconstruct a {\em detail disentangled\/} 3D representation from single-view images. The disentangled details enable detail transfer and 3D reconstruction (shown in two views) with the transferred details from image to another.}
\label{fig:teaser}
\end{figure}

\begin{figure*}
\centering
\includegraphics[width=\textwidth]{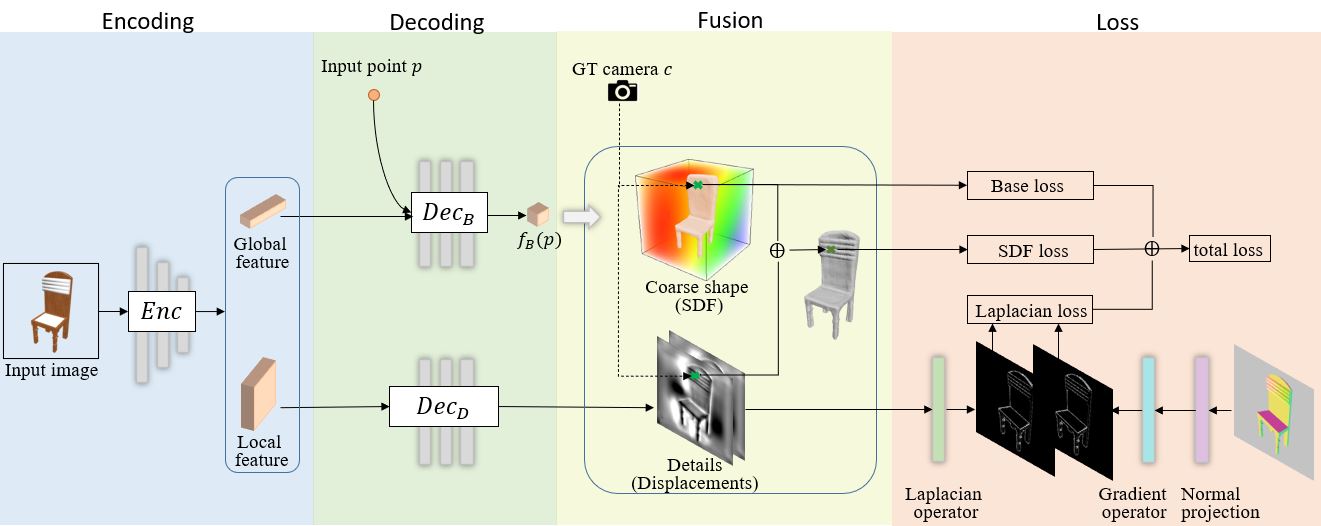}
\caption{The pipeline of our single-view 3D reconstruction network \ddi{} consists of three stages. An encoder extracts global and local features from the input image. This is followed by two decoder branches which respectively predict a base or {\em coarse\/} shape from global features and two displacement maps (back and front) from local features. The final 3D reconstruction is a fusion between the base shape and the displacement maps, with three losses enforcing recovery of coarse shape, overall structure, and surface details (Laplacian).}
\label{fig:network}
\end{figure*}

\begin{figure}
\centering
\includegraphics[width=\linewidth]{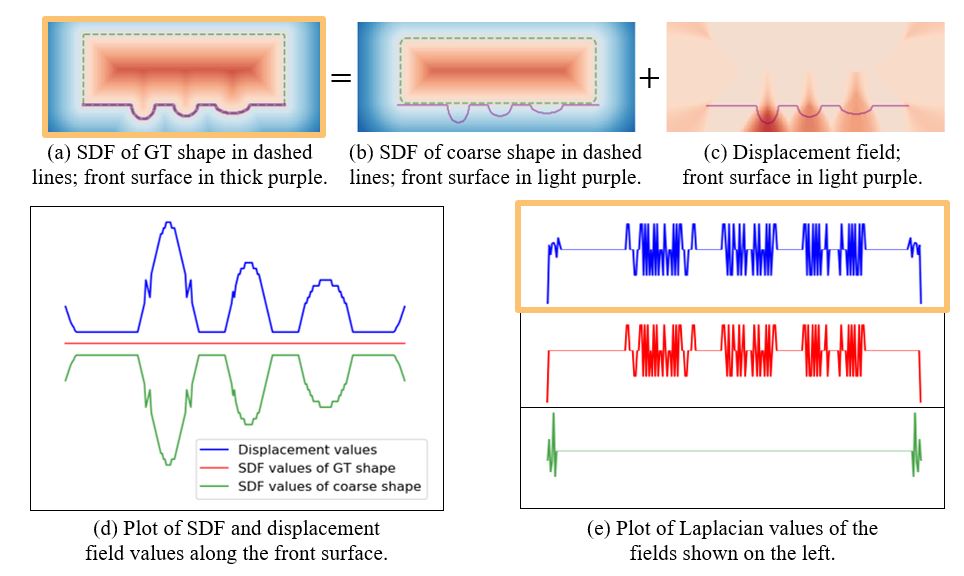}
\caption{Illustration of a ground-truth (GT) shape+SDF (a) and a disentanglement into a base shape+SDF (b) and a displacement field (c). Bottom row plots SDF, displacement field, and Laplacian values along the {\em front surface\/} (\textcolor{purple}{purple} lines) of the GT shape. We see close resemblance between the Laplacian of the displacement field values and that of the GT SDF: \textcolor{blue}{blue} vs.~\textcolor{red}{red} curves in (e). Note that at training, only the GT SDF is known (indicated by \textcolor{orange}{orange} borders in the figure); all other fields are to be {\em learned\/}.}
\label{fig:disentanglement}
\end{figure}

\begin{figure}
\centering
\includegraphics[width=0.98\linewidth]{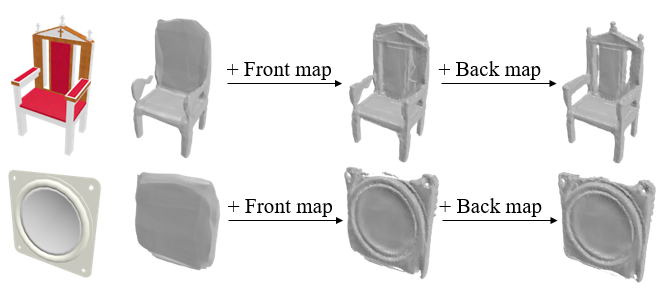}
\caption{A visualization of 3D shapes reconstructed by the two decoders of \ddi{} demonstrates detail disentanglement: our network learns to recover surface details via the front displacement map and other details from the back map. The network was trained on ShapeNet across 13 shape categories. }
\label{fig:twofields}
\end{figure}

Implicit reconstruction networks such as IM-Net~\cite{chen2019learning} and Occupancy Network~\cite{mescheder2019occupancy} learn to predict an implicit function, given a feature encoding of the input image, by minimizing a reconstruction loss. These networks generalize well to new images, but only in terms of the coarse shapes; they are not designed to recover geometric details which are often of small scale and do not incur a sufficient penalty on the loss terms. In a more recent work, DISN, Xu et al.~\cite{xu2019disn} account for both global {\em and local} image features to predict a combined signed distance field (SDF) so as to minimize a {\em single\/} reconstruction loss like prior works. Their network can better resolve structural details, such as the slats in the back of a chair, that are well captured by local image features. However, the rest of the details, in particular {\em surface details\/}, which are just as important for visual perception (e.g., of depth and material), are still not well recovered.

In this paper, we wish to develop an implicit single-view 3D reconstruction network which can recover both topological structures and 
surface details from an input image. Our key idea is that to best reconstruct the details, we ought to train the network to learn a {\em detail disentangled} reconstruction consisting of two functions, one representing the coarse 3D shape and one capturing the details.
However, the main ensuing challenge is that geometric details are so varied that there is no general and reliable way to define what the
details are or what a coarse shape should be. The network must learn the disentangled representations without direct supervision using ground-truth training data.

Figure~\ref{fig:network} illustrates the pipeline of our detail disentangled implicit reconstruction network, coined \ddi{}. Given a single RGB input image, the network encodes it into global and local features which are respectively fed into two decoders. The {\em base decoder\/} uses the global features to reconstruct a coarse (i.e., base) implicit field, while the {\em detail decoder\/} reconstructs, from the local features, a pair of 2D displacement maps, defined over the {\em front\/} and {\em back\/} sides of the captured object that are visible to the camera. 

In the absence of any ground-truth displacement maps for training, or coarse shapes for that matter, we must rely on the original 3D shapes (e.g., from ShapeNet) or their associated SDFs to define the network losses. We first observe that the {\em Laplacian\/} of the SDF of a shape near the shape surface is sensitive to local geometry variations\footnote{The Laplacian of a signed distance function at a point $x$ is proportional to the mean curvature of the isosurface passing through $x$~\cite{esedog2010diffusion}.}, i.e., the surface details. Furthermore, this Laplacian function resembles the Laplacian of the front displacement map if the front side of the coarse shape is mostly flat; see Figure~\ref{fig:disentanglement}. Based on these observations, we define a corresponding {\em Laplacian loss\/} to optimize the front displacement map.

In addition, we define a base loss and an SDF loss, both with respect to the ground-truth SDF, where the SDF loss is computed against a {\em fusion\/} between the predicted coarse SDF and the predicted displacement maps, both the front {\em and the back\/}. As the back displacement map is not factored into the Laplacian loss, it does not capture surface details. However, with local image features as input, the SDF loss does enforce the back map to help reconstruct the overall shape structure, including topological details.
Figure~\ref{fig:twofields} visualizes the disentangled functions our network reconstructs on two examples, where the predicted displacement maps evidently represent shape details, encompassing both topological structures and surface features, while the base decoder reconstructs the coarse shape.

We train our network on ShapeNet Core~\cite{shapenet2015} across all 13 shape categories and show single-view reconstruction results for a variety of 3D objects, as well as reconstruction results from ``images in the wild". We conduct various ablation studies and present both qualitative and quantitative comparisons between \ddi{} and representative single-view 3D reconstruction methods including IM-Net~\cite{chen2019learning} and DISN~\cite{xu2019disn}. While the focus of our work is on reconstructing shape details, evaluations are conducted on images containing objects with varying degrees of geometric details and using different error metrics applicable to overall shapes and edge revelation. 
Finally, we develop and demonstrate a novel application of \ddi{}, where the ability to learn detail functions from images enables
detail transfer from an image onto a reconstructed 3D shape; see Figure~\ref{fig:teaser}.

\section{Related work}
\label{sec:related}

Most learning-based methods for 3D reconstruction aim to generalize to novel data~\cite{tatarchenko2017octree,wu2020pq,wu2017marrnet,richter2018matryoshka,chen2019learning,mescheder2019occupancy,xu2019disn,atlasnet,wang2018pixel2mesh,3DR2N2,wu2018learning,HSP}, while some recent networks are designed to ``overfit'' to specific inputs~\cite{SAL,yariv2020multiview,sitzmann2019siren,mildenhall2020nerf,liu2020neural}.
In the latter case, a network is specifically trained to optimize the reconstruction for a given input, typically multi-view images~\cite{yariv2020multiview,sitzmann2019siren,mildenhall2020nerf,liu2020neural} or a point cloud~\cite{SAL}. As expected, such a specialization tends to produce much higher reconstruction quality compared to methods from the first category. However, with a new input, the network needs to be re-trained. Our work belongs to the first category and in this section, we mainly discuss related works in this category for {\em single-view\/} 3D reconstruction, or SVR, for short.

\vspace{-10pt}

\paragraph{Neural implicit models for SVR.}
The use of deep neural networks for SVR has gained significant improvements with various 3D shape representations, including voxels~\cite{3DR2N2,tatarchenko2017octree,wu2017marrnet,richter2018matryoshka,wu2018learning}, meshes~\cite{wang2018pixel2mesh,atlasnet}, and structural primitives \cite{niu2018im2struct,zou20173d}. 
Recently, implicit representations~\cite{chen2019learning,mescheder2019occupancy,park2019deepsdf,xu2019disn,wu2020pq,michalkiewicz2019deep,littwin2019deep} have emerged as a desirable alternative due to the advantages they offer at representing continuous surfaces with higher visual quality and flexible topology.

Supervised by the ground-truth (GT) occupancy or SDF, earlier implicit reconstruction methods such as IM-NET \cite{chen2019learning}, OccupancyNetwork \cite{mescheder2019occupancy}, and DeepLevelSets \cite{michalkiewicz2019deep} predict the scalar value at each 3D point to approximate the GT. Latent features encoded from the input images are fed into an MLP network together with 3D point coordinates to predict their occupancy or signed distances. Littwin and Wolf~\cite{littwin2019deep} take the encoded feature vectors as the network weights of the MLP to attain a more accurate reconstruction. Instead of predicting the implicit fields as a whole, PQ-NET~\cite{wu2020pq} separately predicts the SDFs for each structural part of the captured object and then combines them together. 

\vspace{-10pt}

\paragraph{Unsupervised SVR.}
Along the lines of SVR without 3D supervision, differentiable renderers~\cite{jiang2020sdfdiff,tulsiani2017multi} have been developed to back-propagate the loss computed from the input images. Liu et al.~\cite{liu2019learning} propose a ray-based field probing technique to render the implicit surfaces to 2D silhouettes, with the geometric details erased from the silhouettes. Niemeyer et al.~\cite{niemeyer2020differentiable} account for both geometry and texture during rendering and make use of rich 2D supervision including RGB, depth, and normal images. 

\vspace{-10pt}

\paragraph{SVR with local image features.}
What is common about {\em all\/} the SVR methods mentioned above is that they are all trained to reconstruct from {\em global\/} image features. As a result, these methods can successfully reconstruct the coarse 3D shapes, but with most shape details missing. A recent work by Tatarchenko et al.~\cite{tatarchenko2019single} reveals that such reconstructions could be easily outperformed by simple retrieval baselines, which may suggest that the main role played by the global images features is recognition rather than reconstruction. This naturally leads to the incorporation of local image features for learning shapes~\cite{jiang2020local,saito2019pifu}. 

Most closely related to our work is DISN~\cite{xu2019disn} which accounts for both global and local image features for SVR. Specifically, it predicts the camera parameters to query the local image feature for each point. Global and local features are processed separately with the point coordinates to obtain two predictions, which are combined and optimized against a {\em single\/} SDF reconstruction loss. In addition, this loss is weighted to place more emphasis on errors associated with small SDF values. Qualitatively, the resulting reconstruction significantly improves the recovery of shape structures, in particular, topological details, but still unable to reconstruct surface details. In a more recent work, LadyBird, Xu et al.~\cite{xu2020ladybird} employ farthest point sampling and feature fusion based on reflective symmetries to deal with self-occlusion. However, geometric details are not taken into account.

Compared to DISN~\cite{xu2019disn}, our network is specifically designed to learn a detail disentangled implicit shape representation, as contrasted in Figure~\ref{fig:twofields}. The key technical difference is that our network defines a dedicated loss for each reconstructed function (the based SDF and two displacement maps) and then sums up the losses, leading to disentanglement, while in DISN, there is only one loss. Specific to the recovery of surface details, we introduce a novel Laplacian loss to learn from GT normal maps.

\vspace{-10pt}

\paragraph{Laplacian-space processing.} 
The Laplacian operator for image or shape processing captures local variations. There have been neural networks which employ Laplacian pyramids to capture multi-scale image structures for coarse-to-fine image generation~\cite{denton2015deep} and super-resolution~\cite{lai2017deep,tang2019deep}. Also, Li et al.~\cite{li2017laplacian} develop a Laplacian loss for neural style transfer to preserve detailed image structures.
However, it is non-trivial to extend Laplacian losses to the SVR framework, where the predicted shape representation must enable the Laplacian computation, while providing alignment to the GT surface. Applying Laplacian losses to surface meshes, as in Pixel2Mesh~\cite{wang2018pixel2mesh}, is more straightforward, e.g., by means of minimizing the error between the predicted Laplacian coordinates before and after mesh deformation. More recently, in ParseNet, Sharma et al.~\cite{sharma2020parsenet} apply the Laplacian loss on parametric surfaces, aligning the GT and the predicted surfaces via Hungarian matching. For implicit methods, existing works such as SoftRas~\cite{liu2019softras} resort to Laplacian regularization to obtain smooth surfaces, rather than detail recovery. In our work, we define disentangled detail functions as displacement maps, which are aligned with the input images, making it possible to define a proper Laplacian loss for SVR with surface details.

\section{Method}
\label{sec:method}

Given a single RGB image of a 3D object, our goal is to reconstruct that object with high-quality shape details, in particular, geometry variations over its {\em surfaces\/}. The input to our reconstruction network consists of the image as well as a 3D point; the network outputs the {\em signed distance\/} from the input point to the target 3D object. Network training is supervised, taking multi-view projections from 3D objects in a shape repository to form the ground-truth data pairs.

Our network learns a {\em disentangled\/} signed distance field (SDF) reconstruction corresponding to the coarse shape and the shape details, employing a novel {\em Laplacian loss\/} to recover surface details. As shown in Figure~\ref{fig:network}, our network starts with an encoder using a CNN architecture to extract image features and two decoders to predict the coarse shape and details separately. The coarse shape and details, both in the form of scalar fields, are then fused together to obtain the SDF of the reconstructed 3D object. 
Finally, we apply Marching Cube~\cite{MC} to extract the zero level set as the final reconstructed 3D output mesh model.

The main challenges include how to disentangle (Section~\ref{disentanglement_subsection}) and how to define the Laplacian loss between network predictions and the ground truth (Section~\ref{loss_subsection}). 

\subsection{Detail disentanglement formulation}
\label{disentanglement_subsection}

The Laplacian of the SDF of a shape near the shape's surface can help detect rapid local geometry variations~\cite{esedog2010diffusion}, i.e., surface details. This motivates the use of Laplacians to help formulate our detail disentanglement under the implicit function setup. 
Specifically, we disentangle the ground-truth SDF $F_{SDF}$ (i.e., the SDF of the ground-truth shape $S$) as the sum of a base implicit field, for a coarse shape, and the residual field which models displacements, as shown along the top of Figure~\ref{fig:disentanglement} and expressed as follows:
\begin{equation}
\begin{split}
& F_{SDF}(p) = f_B(p) + f_D(p), \\
& f_B:\mathbb{R}^3\rightarrow \mathbb{R}, f_D:\mathbb{R}^3\rightarrow \mathbb{R},
\end{split}
\label{eq:disent}
\end{equation}
where $f_B$ and $f_D$ denote the base and displacement fields, respectively, which are learned. 
We follow the convention that capitalization, e.g., $F$, refers to ground-truth functions, while learned functions are given in lower-case.

We assume that the coarse shape is {\em smooth\/} and lies close to the surface $S$. The smoothness herein implies that the (residual) displacement field contains information about surface details. Such information is connected to $F_{SDF}$ through the Laplacian. Furthermore, near $S$, the Laplacian of the displacement field $f_D$ would closely approximate the Laplacian of $F_{SDF}$, if the detail displacements form a height field over a mostly flat surface (on the coarse shape). The latter implies that $\bigtriangleup f_B \approx 0$, hence, due to linearity of the Laplacian operator, we have
\begin{equation}
\begin{split}
\bigtriangleup f_D(p) = 
\bigtriangleup F_{SDF}(p), |dist(p,S)|<\delta.
\end{split}
\label{3D_equation}
\end{equation}
With $|dist(p,S)|<\delta$, only the Laplacian of points near $S$ within a threshold $\delta$ need to be sampled during training.

However, for single-view 3D reconstruction, it is difficult to infer occluded geometry in 3D space. Inspired by recent works \cite{yao2020front2back,saito2020pifuhd} which treat the front and back surfaces separately, our network predicts {\em a pair of 2D displacement maps\/} for the visible front surface and the occluded back surface respectively, instead of a 3D displacement field. The front displacement map recovers details on the visible front surface, by optimizing the Laplacian near that surface against the ground-truth. The back displacement map approximates the residual between the SDF and base distance field to compensate for other details such as topological structures. Putting things together, we have
\begin{equation}
\begin{split}
& F_{SDF}(p)=\left\{\begin{matrix}
f_B(p)+f_{DF}(u(p)), p\in P_F,
\\ 
f_B(p)+f_{DB}(u(p)), otherwise,
\end{matrix}\right. \\
& \bigtriangleup f_{DF}(u(p)) = 
\bigtriangleup F_{SDF}(p), p\in P_F,\\
& f_B:\mathbb{R}^3\rightarrow \mathbb{R}, f_{DF}:\mathbb{R}^2\rightarrow \mathbb{R},
f_{DB}:\mathbb{R}^2\rightarrow \mathbb{R},
\end{split}
\label{2D_equation}
\end{equation}
where $f_{DF}$ and $f_{DB}$ are the displacement maps for the front and back surfaces. $u(p)$ is the operation to project the 3D point $p$ to the pixel position on the image. The point set $P_F$ contains the points near the front surface. 

The advantages of using 2D displacement maps instead of 3D fields are two fold. First, it enables us to learn the small-scale details with contemporary CNN networks. Second, it aligns the details with the input images to compute the Laplacian loss, which we discuss in Section \ref{loss_subsection}.

\subsection{Network pipeline: encoder, decoder, fusion}
\label{network_subsection}

Figure \ref{fig:network} shows the pipeline of our network \ddi{}. The encoding uses a CNN to extract the global feature vector and local feature map from the input image. 
The base decoder $Dec_B$, an MLP, takes the global feature vector with a 3D point coordinate as input, and outputs the base value of this point, i.e., the signed distance from this point to the coarse shape. The detail decoder $Dec_{D}$ contains the residual convolutional layers with the local feature map as input, and outputs a front displacement map encoding surface details on the visible front surface, and a back map to compensate for the topology details on the back surfaces. The back displacement map is necessary since a pixel outside the object mask should affect all the points along the ray. 

The third stage is to fuse the base distance field with two displacement maps. Similar to DISN, we train a separate network to predict the camera parameters to query the displacement values per point on the displacement maps. As in equation (\ref{2D_equation}), the base distance of a point is summed up with its corresponding value queried from the front displacement map, if this point is closer to the visible front surface. Otherwise, we sum the base distance and the corresponding value from the back displacement map. In the implementation, we simply estimate the gradient of the SDF at each point with central difference approximation. If the gradient direction is close to the viewpoint direction and the ground-truth SDF is smaller than a threshold, we classify the point as near the front surface. Note that we use the ground-truth camera parameters and the gradients estimated from ground-truth SDF during training, and the predictions during testing. 

\subsection{Network losses}
\label{loss_subsection}

Our loss function is formulated as $L = L_{B} + L_{lap} + L_{sdf}$, where $L_{B}$, $L_{lap}$, and $L_{sdf}$ denote the base loss, Laplacian loss, and SDF loss, respectively. Specifically, $L_{B}$ is the L2-distance between the predicted base distance field $f_B$ and the ground-truth SDF $F_{SDF}$ over a set of sample points to learn the coarse shape. The SDF loss term $L_{sdf}$ is the L1-distance between the fused implicit field $f$ and the ground-truth SDF $F_{SDF}$; this term serves as a regularization for the displacement maps. Thus we have,
\begin{equation}
\begin{split}
& L_{B}=\frac{1}{M} \sum_{i=1}^{M}\left \|f_B(p_i)-F_{SDF}(p_i)\right \|^2_2 
\\
& L_{sdf}=\frac{1}{M} \sum_{i=1}^{M}|f(p_i)-F_{SDF}(p_i)|
\end{split}
\label{loss_function}
\end{equation}
The Laplacian loss, $L_{lap}$, aims to minimize the error between the Laplacian of the predicted (front) displacements and the Laplacian of the ground-truth SDF. However, there exists a mismatch between the two Laplacians since our disentangled details are displacement maps defined in 2D while the ground-truth SDFs are defined in 3D.

To solve this problem, we estimate the {\em 2D projection\/} of the ground-truth Laplacian, i.e., the Laplacian of the ground-truth SDF with respect to pixel positions on the image. This is reasonable since the single-view images are not sensitive to variations along the viewing direction. In addition, this enables us to obtain the ground-truth Laplacian from 2D normal maps, instead of computing it in 3D.

To project a point $p$ in 3D space, we first transform it to $p'=(p'_x,p'_y,p'_z)$ in the camera's viewpoint, and then project it to the pixel position $u(p)=(u_x,u_y)$. The Laplacian of the front displacement map is 
\begin{equation}
\bigtriangleup f_{DF}(u(p))=\frac{\partial ^2 f_{DF}(u(p))}{\partial (u_x)^2} + \frac{\partial ^2 f_{DF}(u(p))}{\partial (u_y)^2}.
\end{equation}
If $p$ lies on the visible front surface, the ground-truth normal map provides its unit normal vector $N(u(p))=\frac{\partial F_{SDF}(p)}{\partial p'}$, which equals to the gradient of the SDF with respect to the point coordinates $p'$ in the camera view. With the camera parameters in the projecting operation, we obtain the gradient of the coordinates $p'$ with respect to the pixel position $u(p)$, denoted by $\frac{\partial p'}{\partial u(p)}$. Therefore, we have the projected gradient of the SDF with respect to $u(p)$ as 
\begin{equation}
N'(u(p))=(N(u(p))\cdot\frac{\partial p'}{\partial u_x}, N(u(p))\cdot\frac{\partial p'}{\partial u_y}),
\end{equation}
and the projected Laplacian (the ground-truth Laplacian) is
\begin{equation}
l(u(p))=N(u(p))\cdot\frac{\partial p'}{\partial ^2 u_x}+ N(u(p))\cdot\frac{\partial p'}{\partial ^2 u_y}.
\end{equation}
Hence, the Laplacian loss is defined as
\begin{equation}\label{eq:lap_loss}
L_{lap}=\frac{1}{|P_F|} \sum_{p_i \in P_F}\left \|\bigtriangleup f_{DF}(u(p_i))-l(u(p_i))\right \|^2_2.\\
\end{equation}

\noindent \textbf{Weighted sampling.} The loss terms are all defined on a set of sampled points. Unlike previous works, e.g.,~\cite{chen2019learning,xu2019disn}, which randomly sample near object surfaces, we emphasize the importance of small-scale (e.g., thin) structures. 
Assuming a dense set of point-value pairs for an object, we define the density at each point as the number of points in its neighborhood with a prescribed radius. The interior points only count their neighbor points inside the object, so do the exterior points. During training, we sample an equal number of interior and exterior points with their densities as sampling weights. Such a weighted sampling strategy enables us to have more interior point-value pairs for the thin structures to better recover them during reconstruction.

\section{Results, evaluation, and application}
\label{sec:results}

All the reconstruction networks are trained (over all categories) and tested on the ShapeNet Core dataset \cite{shapenet2015}. The training set comes from the ground-truth SDFs provided by DISN \cite{xu2019disn} and their rendered images including single-view images and 2D normal maps. The ground-truth SDFs were randomly sampled on 32,768 points near the object surfaces with their signed distance values. In each iteration during training, we randomly select 2,048 points with our weighted sampling strategy to compute the loss and update the network. For the input images, we use all the views during training but use the one showing most of the details during testing to better evaluate the shape details.

\subsection{Evaluation metrics}
\label{evalutaion_metrics}

For all the implicit 3D reconstructions we test, the final meshes are extracted via MarchingCubes in $128^3$ resolution. To measure the overall reconstruction quality, we use Chamfer $L_1$ Distance (CD) \cite{mescheder2019occupancy} with 20K sampled points and Intersection of Union (IoU) in $32^3$ resolution. It is worth noting however that despite their popularity, CD and IoU are not the best measures of {\em visual\/} reconstruction quality~\cite{jin2020drkfs}. Also, they do not emphasize on small-scale details.

Since the focus of our work is on detail recovery, we employ the Edge Chamfer Distance (ECD) \cite{chen2020bspnet}, which is defined as the CD between the edge points on the ground-truth shapes and the reconstructions. The ``edgeness'' of each point $p_i$ is estimated as $\sigma (p_i)=min_{p_j \in \mathcal{N}_i}|n_i \cdot n_j|$, where $\mathcal{N}_i$ contains neighbors of point $p_i$, $n_i$ and $n_j$ are the unit normal vectors for points $p_i$ and $p_j$. From 20K sampled points, we retrieve the nearest 10 neighbors for each point and retain the points 
with $\sigma (p_i)<0.8$ to measure the small-scale details. Similarly, we develop a 2D version of the ECD metric, since we recover details observed from images. ECD-2D is defined as the CD between the edge pixels on the corresponding renderings. We apply the Canny edge detector \cite{canny1986computational} on the rendered $224\times224$ normal map of the reconstructed objects to obtain the edge pixels. The original ECD and its 2D version are denoted as ECD-3D and ECD-2D in our quantitative evaluation. 

\subsection{Ablation study}
\label{ablation_study}

\begin{figure}
\centering
\includegraphics[width=0.99\linewidth]{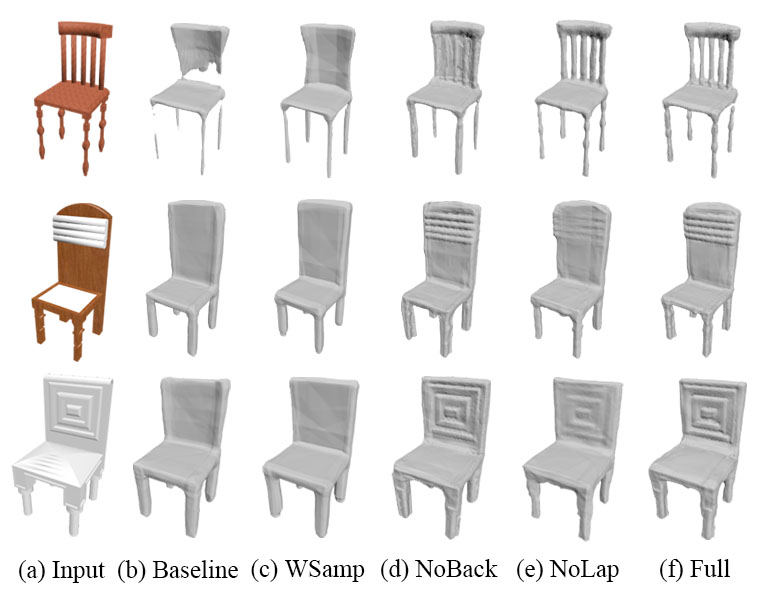}
\caption{Qualitative results for ablation study: reconstructed objects rendered with the same camera parameters as input images.}
\label{fig:ablation_quality}
\end{figure}

\begin{table}
\begin{center}
\setlength{\tabcolsep}{2.35mm}{
\begin{tabular}{ |l|c|c|c|c|c|c| } 
 \hline
  & CD & IoU & ECD-3D & ECD-2D \\ 
 \hline
 Baseline & 0.0417 & 0.523 & 0.0735 & 3.304 \\ 
 WSamp & 0.0340 & 0.587 & 0.0624 & 2.626 \\ 
 NoBack & 0.0306 & 0.589 & 0.0525 & 1.802 \\
 NoLap & 0.0302 & 0.601 & 0.0524 & 1.653 \\
 Full & \textbf{0.0297} & \textbf{0.613} & \textbf{0.0503} & \textbf{1.456} \\ 
 \hline
\end{tabular}
\caption{Quantitative evaluation for ablation study.}
\label{table:ablation_quantity}
}
\end{center}
\end{table}

We conduct an ablation study to show how each component of \ddi{} contributes to detailed single-view 3D reconstruction. For the study, the
networks are trained on the chair category from ShapeNet with the ground-truth camera parameters assumed given. The network options are:
\begin{itemize}
\item {\em Baseline\/}: no detail decoder from \ddi{} and trained with uniform sampling and with loss $L_{sdf}$ defined on the output of the base decoder. 
\vspace{-5pt}
\item {\em WSamp\/}: weighted sampling to train the baseline.
\vspace{-5pt}
\item {\em NoBack\/}: no back displacement map prediction from \ddi{}; the predicted base distance field is fused with only the front displacement map. 
\vspace{-5pt}
\item {\em NoLap\/}: only removing $L_{lap}$ loss from \ddi{}; both NoBack and NoLap use weighted sampling. 
\vspace{-5pt}
\item {\em Full\/}: all-component \ddi{} as describe in Figure~\ref{fig:network}.
\end{itemize}

Figure \ref{fig:ablation_quality} and Table \ref{table:ablation_quantity} provide qualitative and quantitative comparison results, respectively. As we can see, weighted sampling helps reconstruct thin volumes, with the detail decoder providing even more improved results on topological structures, while surface details are best recovered with the Laplacian loss (see NoLap vs.~Full or NoBack).

\subsection{Comparison}
\label{comparison}

\begin{figure*}
\centering
\includegraphics[width=0.99\linewidth]{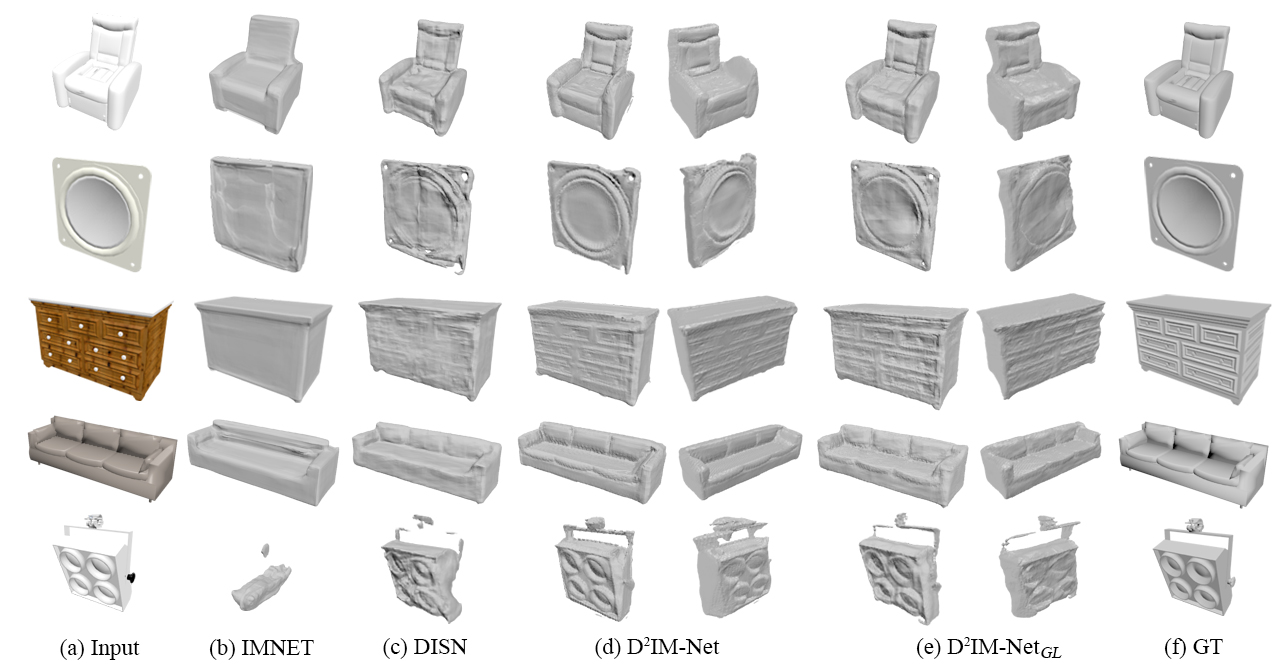}
\caption{Qualitative comparison between reconstruction results by IMNET \cite{chen2019learning}, DISN \cite{xu2019disn}, \ddi{}, and \ddg{}.}
\label{fig:comparison_quality}
\end{figure*}
\begin{table*}
\centering
\resizebox{\textwidth}{!}{
\begin{tabular}{ |c|l|ccccccccccccc|c| } 
 \hline
  & & plane & bench & box & car & chair & display & lamp & speaker & rifle & sofa & table & phone & boat & Mean \\ 
 \hline
 
  IoU $\uparrow$ & IMNET & 0.5200 & 0.5133 & 0.4581 & 0.7653 & 0.5411 & {\em 0.5185\/} & 0.4168 & {\bf 0.5194\/} & 0.5643 & 0.6386 & 0.5083 & 0.6701 & 0.5631 & 0.5536 \\
 & DISN & 0.5362 & 0.5403 & 0.4615 & {\em 0.8105\/} & {\em 0.5539\/} & 0.4879 & 0.3791 & 0.4958 & \textbf{0.7237} & {\em 0.6520\/} & \textbf{0.5629} & {\em 0.7071\/} & \textbf{0.6566} & 0.5821 \\
 & \ddi{} & \textbf{0.5584} & \textbf{0.5495} & \textbf{0.4860} & 0.7980 & \textbf{0.5613} & \textbf{0.5272} & {\em 0.4213\/} & {\em 0.5175\/} & {\em 0.6813\/} & \textbf{0.6535} & 0.5367 & \textbf{0.7616} & 0.6339 & \textbf{0.5912} \\
 & \ddg{} & {\em 0.5553\/} & {\em 0.5425\/} & {\em 0.4760\/} & \textbf{0.8114} & 0.5441 & 0.5112 & \textbf{0.4495} & 0.5031 & 0.6626 & 0.6437 & {\em 0.5475\/} & 0.6966 & {\em 0.6381\/} & {\em 0.5832\/} \\
 
 \hline
 CD $\downarrow$ & IMNET & 0.0426 & 0.0382 & 0.0503 & 0.0437 & 0.0376 & 0.0479 & {\em 0.0557\/} & 0.0632 & 0.0329 & 0.0475 & 0.0432 & 0.0317 & 0.0443 & 0.0445 \\
 & DISN & 0.0398 & 0.0351 & 0.0412 & \textbf{0.0308} & {\em 0.0326\/} & 0.0462 & 0.0770 & 0.0647 & \textbf{0.0199} & \textbf{0.0366} & {\em 0.0316\/} & 0.0282 & \textbf{0.0312} & 0.0396  \\
 & \ddi{} & \textbf{0.0358} & \textbf{0.0312} & \textbf{0.0385} & 0.0348 & 0.0329 & \textbf{0.0422} & {\em 0.0557\/} & \textbf{0.0561} & 0.0244 & 0.0391 & 0.0356 & \textbf{0.0245} & {\em 0.0339\/} & {\em 0.0373\/} \\
 & \ddg{} & \textbf{0.0358} & {\em 0.0337\/} & {\em 0.0386\/} & {\em 0.0313\/} & \textbf{0.0308} & {\em 0.0427\/} & \textbf{0.0549} & {\em 0.0572\/} & {\em 0.0242\/} & {\em 0.0375\/} & \textbf{0.0310} & {\em 0.0270\/} & {\em 0.0339\/} & \textbf{0.0368} \\
 
 \hline
 ECD-3D $\downarrow$ & IMNET & 0.0789 & 0.0685 & 0.0872 & 0.0872 & 0.0661 & 0.0820 & 0.0995 & 0.1080 & 0.0674 & 0.0790 & 0.0710 & 0.0724 & 0.0823 & 0.0807 \\
 & DISN & 0.0684 & 0.0573 & 0.0697 & {\em 0.0680\/} & 0.0564 & 0.0765 & 0.1127 & 0.1077 & {\em 0.0350\/} & {\em 0.0606\/} & {\em 0.0601\/} & 0.0708 & 0.0583 & 0.0694 \\
 & \ddi{} & \textbf{0.0567} & \textbf{0.0477} & \textbf{0.0661} & 0.0728 & {\em 0.0523\/} & \textbf{0.0674} & {\em 0.0918\/} & \textbf{0.0909} & \textbf{0.0343} & 0.0642 & 0.0630 & \textbf{0.0609} & {\em 0.0568\/} & \textbf{0.0634} \\
 & \ddg{} & {\em 0.0598\/} & {\em 0.0516\/} & {\em 0.0691\/} & \textbf{0.0646} & \textbf{0.0504} & {\em 0.0713\/} & \textbf{0.0897} & {\em 0.0973\/} & 0.0357 & \textbf{0.0602} & \textbf{0.0567} & {\em 0.0660\/} & \textbf{0.0534} & {\em 0.0635\/} \\
 
 \hline
 ECD-2D $\downarrow$ & IMNET & 2.532 & 2.845 & 4.467 & 3.344 & 2.703 & 3.230 & 3.361 & 4.198 & 3.138 & 2.979 & 2.846 & 2.422 & 3.046 & 3.162 \\
 & DISN & 2.672 & 2.209 & 2.250 & {\em 2.042\/} & 1.983 & 3.156 & 4.863 & 3.338 & {\em 1.353\/} & 2.062 & 2.065 & 2.259 & {\em 2.003\/} & 2.481 \\
 & \ddi{} & {\em 1.991\/} & \textbf{1.666} & {\em 1.794\/} & 2.072 & {\em 1.707\/} & \textbf{1.954} & {\em 3.157\/} & \textbf{2.636} & \textbf{1.277} & {\em 2.014\/} & {\em 1.880\/} & \textbf{1.617} & \textbf{1.730} & \textbf{1.961} \\
 & \ddg{} & \textbf{1.982} & {\em 1.774\/} & \textbf{1.739} & \textbf{1.767} & \textbf{1.584} & {\em 2.675\/} & \textbf{3.009} & {\em 2.715\/} & 1.766 & \textbf{1.776} & \textbf{1.737} & {\em 2.142\/} & 2.269 & {\em 2.072\/} \\
 
 \hline
\end{tabular}}
\caption{Quantitative comparison results: IoU at $32^3$ resolution; CD and ECD-3D on 20K sample points; ECD-2D on $224\times224$ rendered normal maps. Top numbers are in bold and second place is indicated in italic.}
\label{table:comparison}
\end{table*}

\begin{figure*}
\centering
\includegraphics[width=0.99\linewidth]{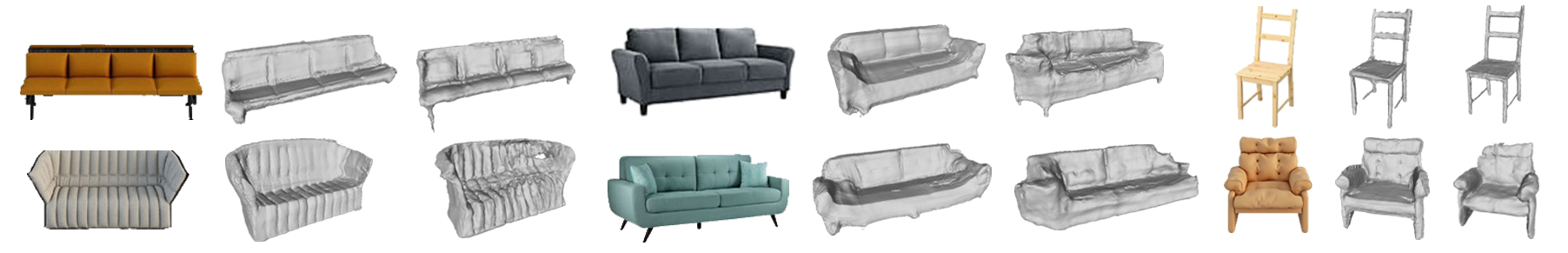}
\caption{Reconstruction results from single-view images ``in the wild'' using \ddi{} (left) and \ddg{} (right). }
\label{fig:online_product_recon}
\end{figure*}

In our comparison to the state of the art, we focus on implicit models which have yielded the best reconstruction quality so far. In addition to IMNET \cite{chen2019learning}, which is a baseline corresponding to the base decoder branch of \ddi{}, we focus on comparing to DISN \cite{xu2019disn}, which is, to the best of our knowledge, the top single-view reconstruction network to date in terms of detail recovery. We also test a slight variant to \ddi{}, called \ddg{}, which takes both global and local features as input to its base decoder.

As shown in Figure \ref{fig:comparison_quality}, IMNET generally obtains good coarse reconstruction, but misses most details. DISN does a better job in terms of recovering topological structures and shape boundaries, but typically blurs surface features. Both versions of \ddi{} visually outperform IMNET and DISN, especially over small-scale, high-frequency details. This is consistent with the quantitative results, provided by ECD-3D and ECD-2D measures, as shown in Table \ref{table:comparison}. Overall, Table \ref{table:comparison} shows that both versions of \ddi{} also outperform
IMNET and DISN quantitatively, in terms of both overall reconstruction quality (CD and IoU) and edge feature recovery (ECD-3D and ECD-2D). 

Comparing between \ddi{} and \ddg{}, we generally find \ddi{} to slightly outperform the latter in visual quality (see Figure \ref{fig:comparison_quality}), especially in terms of surface details, which may be due to the redundancy in using latent (local) features in both the base and detail decoders by \ddg{}. \ddg{} appears to perform better on thin structures. Results from Figure \ref{fig:online_product_recon} support these findings, where we show single-view 3D reconstruction from several online images, with no 3D ground-truth shapes.
\subsection{Application: detail transfer and reconstruction}

\begin{figure}
\centering
\includegraphics[width=0.99\linewidth]{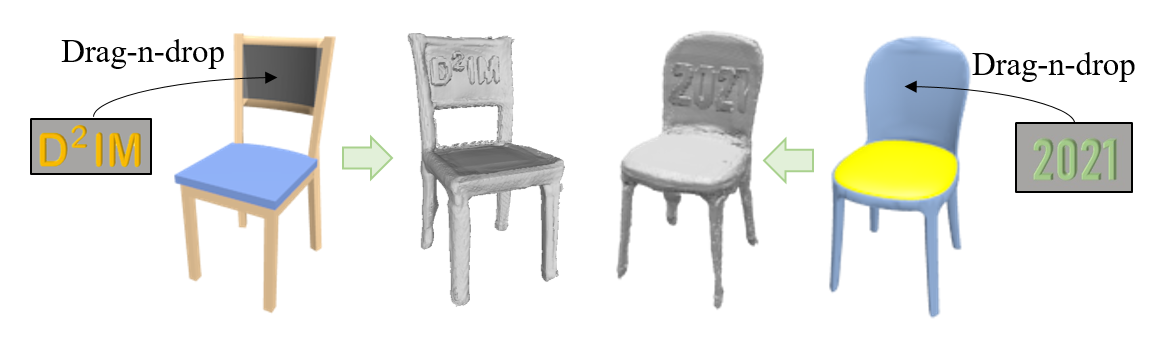}
\caption{
When a logo image, e.g., of ``$D^2IM$'', is ``drag-n-dropped" onto a chair image, we obtain a reconstructed 3D chair model (shown in a view that is different from that of the input image) with surface features resembling the input logo.} 
\label{image_transfer}
\end{figure}

With disentangled coarse shapes and details in the context of 3D reconstruction, enabled by our work, it becomes possible to {\em transfer\/} geometric details or features from images to images and then obtain a final 3D outcome. 

\vspace{-10pt}

\paragraph{Detail transfer.} 
Given a pair of single-view images of different objects (e.g., two chairs), our network predicts their disentangled coarse shapes and details, respectively. Detail transfer involves fusing the disentangled {\em source\/} details with the {\em target\/} coarse shape. In the fusion stage, for each point $p$, we sum up its base distance $f_B(p)$ from the target image and the queried source detail displacement $f_{DF}(u_S(q))$ or $f_{DB}(u_S(q))$ with a learned 3D correspondence $q=C_{T \rightarrow S}(p)$, where $u_S$ is the projection operation with camera parameters predicted from the source image. 

Our method allows such a detail transfer for a specified semantic part, by fusing the displacement values from the source image for the points near this part, and displacement values from the target image otherwise. Results in Figure \ref{fig:teaser} show surface detail transfer from the source chair images (top row) to the target chair images (left column) on the chairs' backs while preserving the coarse shapes.

In the implementation, we use a pre-trained semantic segmentation network \cite{dgcnn} on the two coarse shapes to build the correspondence $q=C_{T \rightarrow S}(p)$. The corresponding segmented parts imply a point-wise correspondence within the local volumes. Specifically, for each point $p$, we compute its local coordinates with respect to the frame defined by the target part it belongs to, and then map it back to the world coordinates $q$ based on the frame of the source part. The local frames are origined at the center of the axis-aligned bounding boxes of each part with fixed axes directions.

\vspace{-10pt}

\paragraph{``Paste-n-reconstruct''.}
Under the same spirit of image-to-image detail transfer but in a slightly different task setting, Figure \ref{image_transfer} shows how a small image logo can be drag-n-dropped onto another image, where the logo content is pasted onto the target image and then a 3D shape can be reconstructed with the pasted logo features.

To implement this, the target image (the chair in Figure~\ref{image_transfer}) goes through the \ddi{} encoder and base decoder to provide the base distance field for the coarse shape. On the other hand, both the target image and the (source) logo image go through the same encoder and detail decoder to predict their displacement maps. With the separately predicted (or pre-defined) camera parameters for each image, we fuse the base distance field and all the displacement maps (only front displacement maps of the logo images) by the projection with their camera parameters. When the foreground masks of the logo images are given, we can crop the foreground displacements for a better visualization.

\section{Conclusion, limitation, and future work}
\label{sec:future}

We tackle perhaps the ``last mile'' in single-view 3D reconstruction, i.e., to recover small-scale geometric details, especially
surface features. This is a deceptively difficult problem as we seek a network that generalizes to shapes across multiple 
categories (13 categories in ShapeNet in our experiments), not a method that ``overfits" to specific inputs. Note also that
we do not rely on symmetry priors or color/material cues. Our key idea is to learn a detail disentangled representation with
a dedicated loss for surface details, defined in the Laplacian domain.
 
One main limitation of our current method is the assumption that the surface details are defined by a height field 
over a mostly flat surface. One implication of this is that geometric details corresponding to ``overhangs'' are precluded. 
Another implication is that, technically, our network would be unable to recover surface details over surfaces that are 
sufficiently curved. In practice, we have found that our network is able to recover surface details over mildly curved 
surfaces, as the example at the bottom-left of Figure~\ref{fig:online_product_recon} demonstrates.
A second limitation is that our Laplacian loss is defined only on the front surface of the recovered
shape. Furthermore, even on the front, we can notice that the reconstructions obtained often look slightly worse when viewed from
an different angle as in the input image. Possible remedies to this include more accurate view parameter inference and
consideration of symmetry priors~\cite{yao2020front2back}.

In addition to addressing the above limitations, we are also interested in expanding the use of neural Laplacian domain 
processing to other shape representations such as voxels, point clouds, and meshes, as well as exploring disentangled 
learning of geometric details for a variety of other applications including multi-modal detail transfer, 3D superresolution,
and generative shape modeling.

{\small
\bibliographystyle{ieee_fullname}
\bibliography{references}
}

\end{document}


\title{\ddi: Learning Detail Disentangled Implicit Fields from Single Images}

\author{
Manyi Li
\qquad
Hao Zhang\\ 
Simon Fraser University
}

\maketitle
\section{Implementation}

In the implementation of \ddi{}, we take ResNet18 as our encoder to obtain the global feature and local feature map from the input image. The base decoder is an MLP with the architecture of IMNET \cite{chen2019learning}. The detail decoder follows the network in \cite{YuZLZG19} to predict the two displacement maps from the local feature map. As for \ddg{}, we take DISN \cite{xu2019disn} as the base decoder with both their global decoder and local decoder. 

In the Laplacian computation, in order to balance the three loss terms, we scale the predicted and ground-truth derivitaves by the same factor with respect to $\frac{\partial u(p)}{\partial p'}$. 
Therefore, the Laplacian loss becomes (see Section 3.3 in the main paper for the denotations)
\begin{equation}
\begin{split}
\label{eq:lap_loss}
&L_{lap}=\frac{1}{|P_F|} \sum_{p_i \in P_F}\left \|\bigtriangleup f'_{DF}(u(p_i))-l'(u(p_i))\right \|^2_2\\
&l'(u(p))=\frac{N(u(p))}{\partial u_x}+\frac{N(u(p))}{\partial u_y}\\
&\bigtriangleup f'_{DF}(u(p))=\frac{f_{DF}(u(p))}{\partial ^2 u_x}  \cdot \frac{\partial u_x}{\partial p'_x} + \frac{f_{DF}(u(p))}{\partial ^2 u_y}  \cdot \frac{\partial u_y}{\partial p'_y}.
\end{split}
\end{equation}

\section{Detail transfer results}
We present more results of detail transfer between two images. The details on the chairs' backs are transferred from the source images to the target images. Both the source images and target images are from the test set.

\begin{figure}[htbp]
\centering
\includegraphics[width=0.8\linewidth]{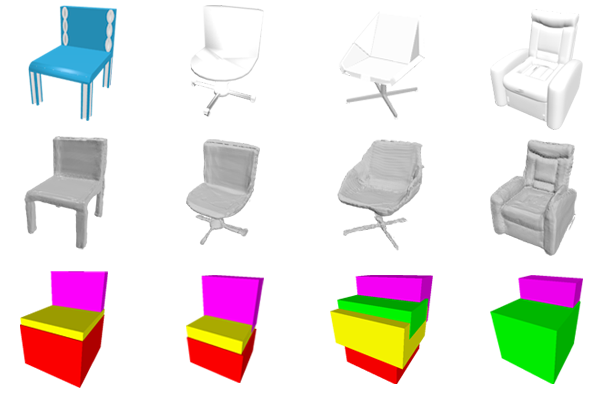}
\caption{The reconstructions (middle row) and part segmentation (bottom row) of the input images (top row). The images are used as the target images in Figure \ref{transfer}.}
\label{segmentation}
\end{figure}

\begin{figure*}
\centering
\includegraphics[width=\textwidth]{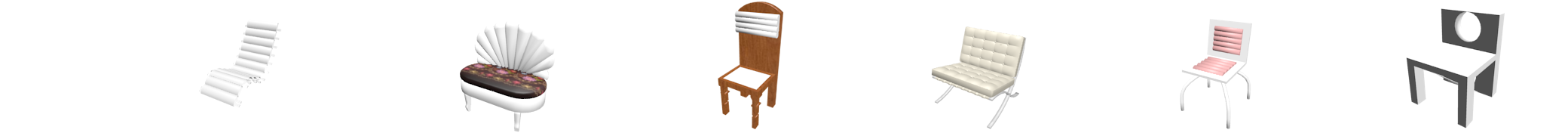}
\includegraphics[width=\textwidth]{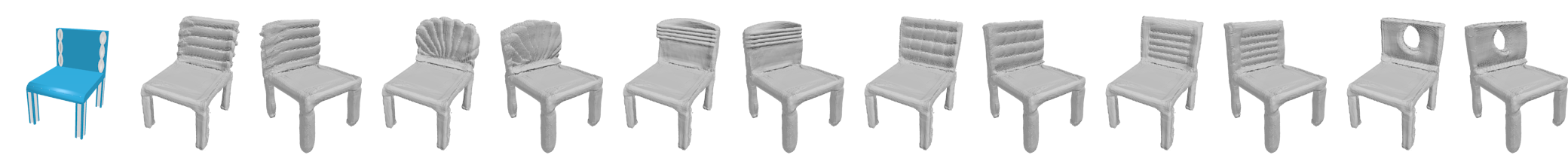}
\includegraphics[width=\textwidth]{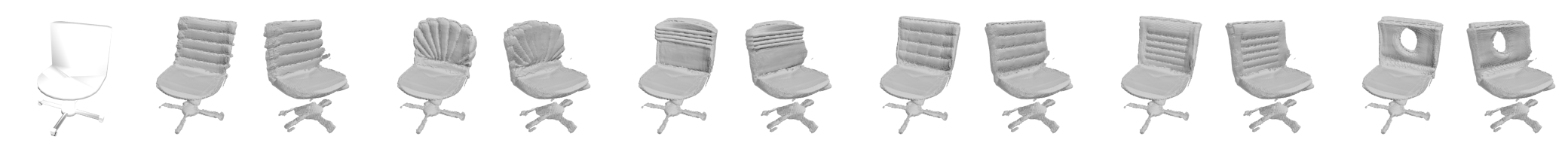}
\includegraphics[width=\textwidth]{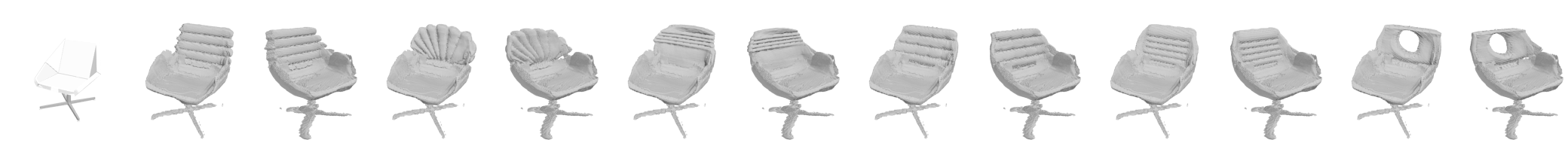}
\includegraphics[width=\textwidth]{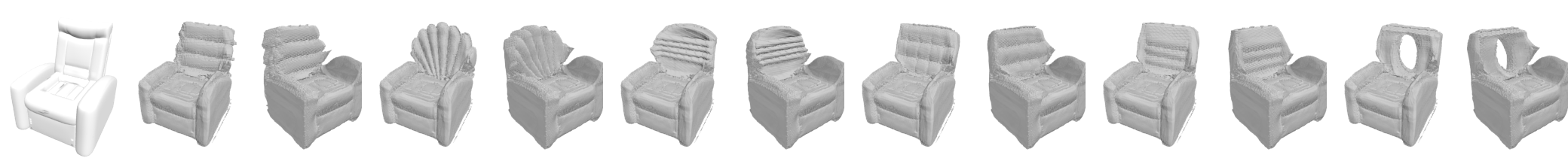}

\caption{More results of detail transfer between semantic parts (chairs' backs). Top row: source image to provide details; Left column: target image to provide coarse shapes. Two views of each transferred reconstruction are shown.}
\label{transfer}
\end{figure*}

For the target images, we show the reconstructions and their part segmentation (axis-aligned bounding box per part) \cite{dgcnn} in Figure \ref{segmentation}, the detail transfer results in Figure \ref{transfer}. As described in Section 4.4 of our paper, the semantic segmentation of the reconstructed coarse shapes are used to provide the 3D correspondence for the transfer. Note that one can also interactively tune the bounding boxes to refine the transferred details.

\section{Single-view reconstruction results}
\begin{figure*}
\centering
\includegraphics[width=\textwidth]{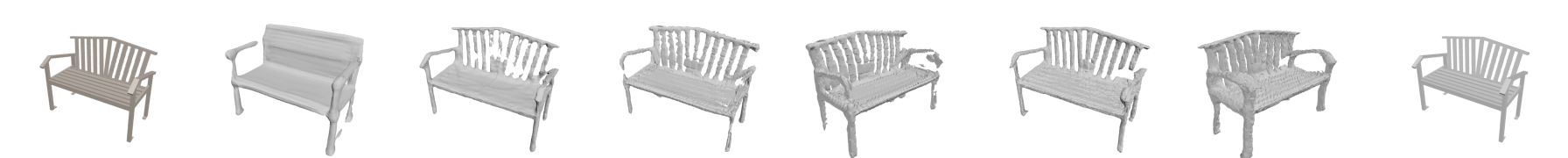}
\includegraphics[width=\textwidth]{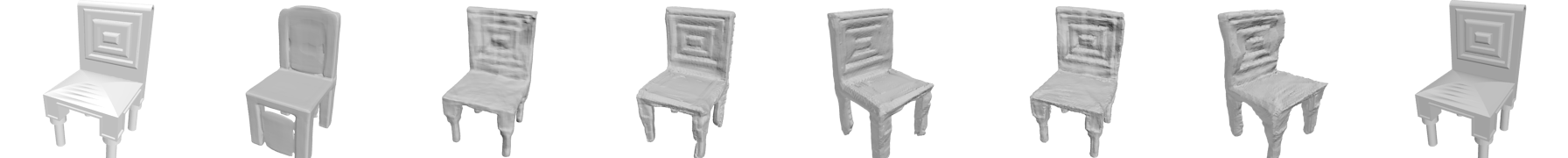}
\includegraphics[width=\textwidth]{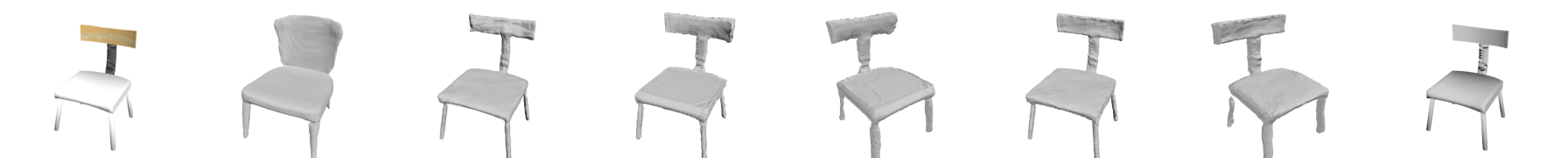}
\includegraphics[width=\textwidth]{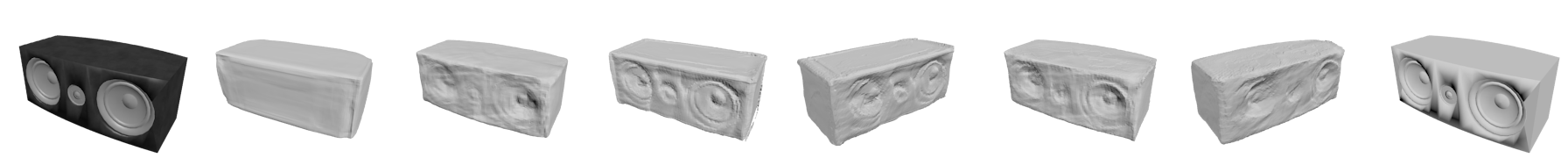}
\includegraphics[width=\textwidth]{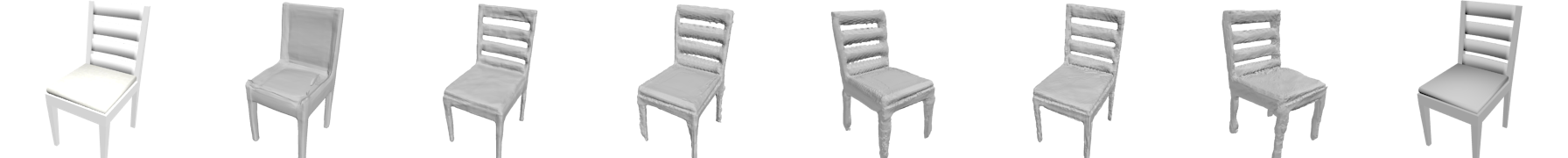}
\includegraphics[width=\textwidth]{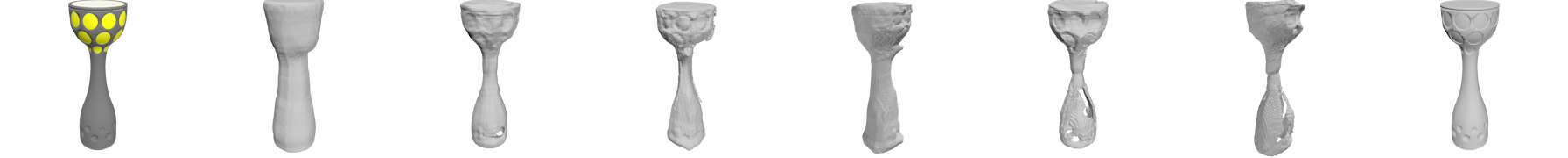}
\includegraphics[width=\textwidth]{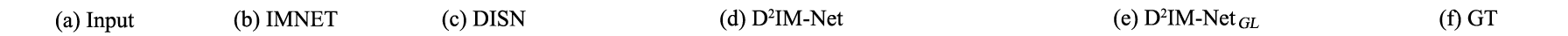}
\caption{More qualitative results. Two views of \ddi{} and \ddg{} are presented to show the reconstruction and recovered details.}
\label{comparison1}
\end{figure*}

\begin{figure*}
\centering
\includegraphics[width=\textwidth]{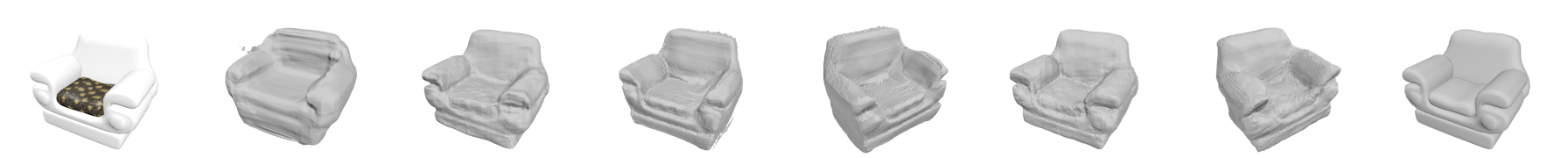}
\includegraphics[width=\textwidth]{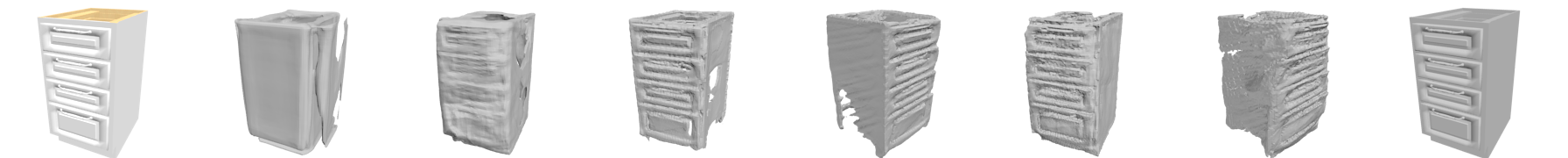}
\includegraphics[width=\textwidth]{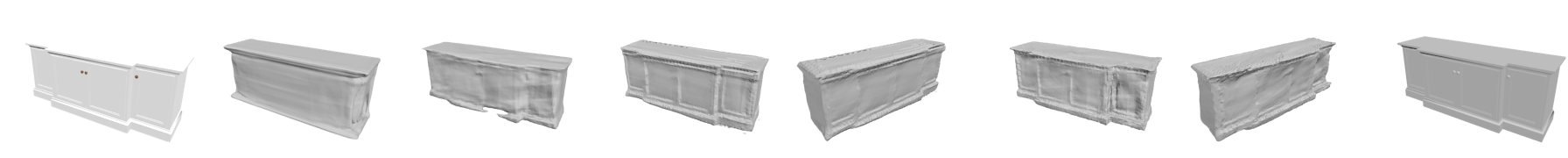}
\includegraphics[width=\textwidth]{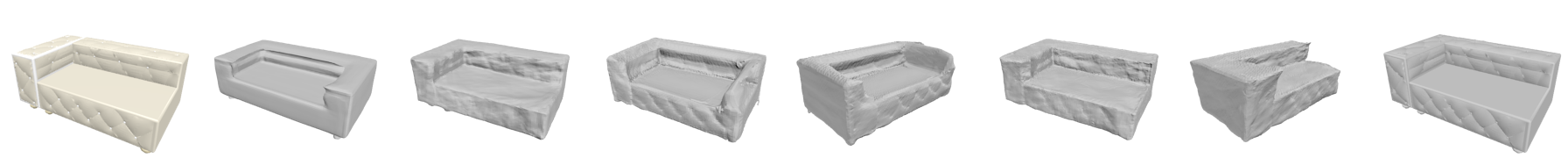}
\includegraphics[width=\textwidth]{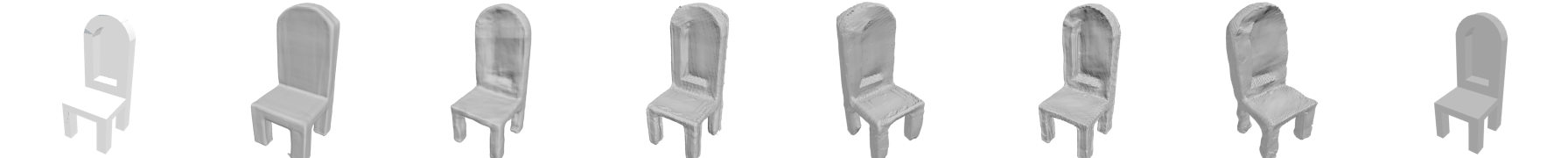}
\includegraphics[width=\textwidth]{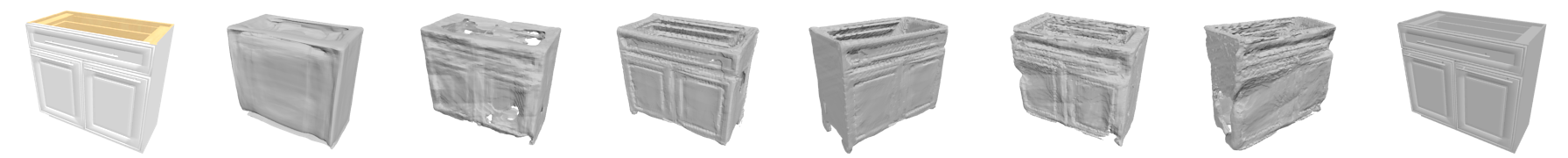}
\includegraphics[width=\textwidth]{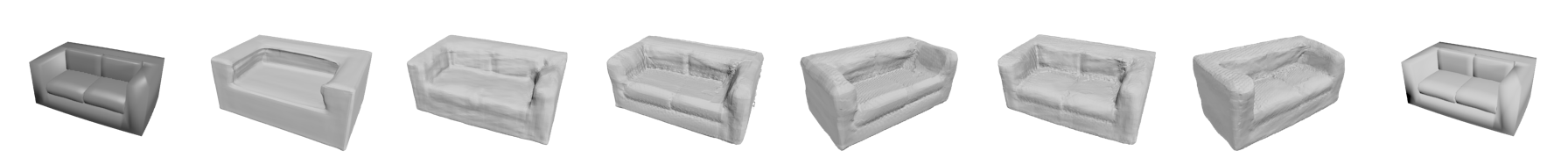}
\includegraphics[width=\textwidth]{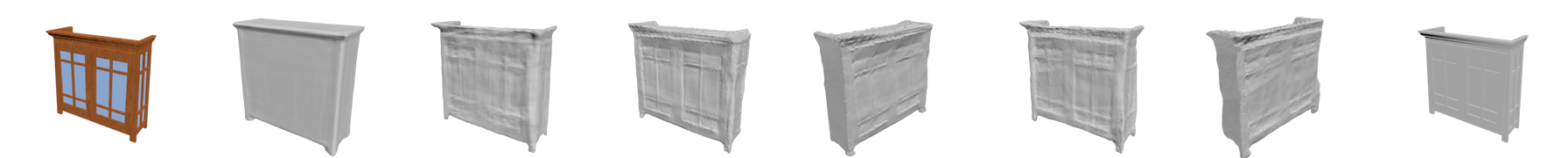}
\includegraphics[width=\textwidth]{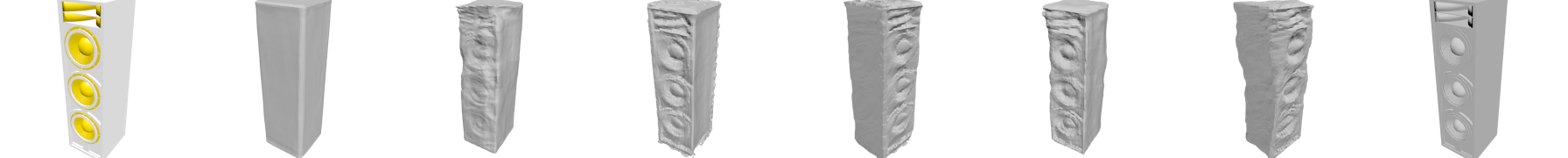}
\includegraphics[width=\textwidth]{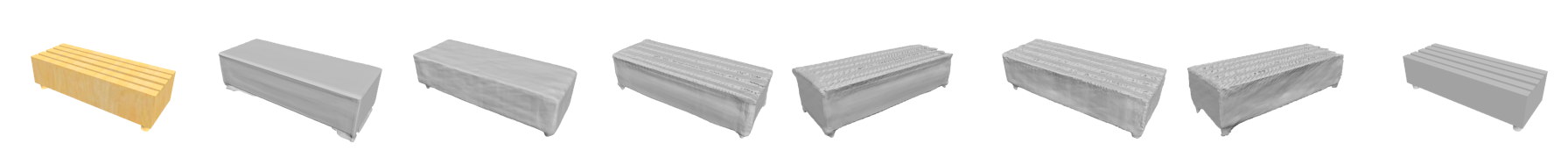}
\includegraphics[width=\textwidth]{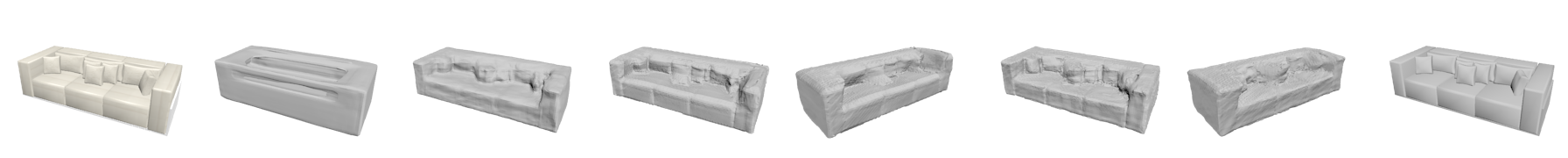}
\includegraphics[width=\textwidth]{images/supp_images/text.png}
\caption{More qualitative results. Two views of \ddi{} and \ddg{} are presented to show the reconstruction and recovered details.}
\label{comparison2}
\end{figure*}

\begin{figure*}
\centering
\includegraphics[width=\textwidth]{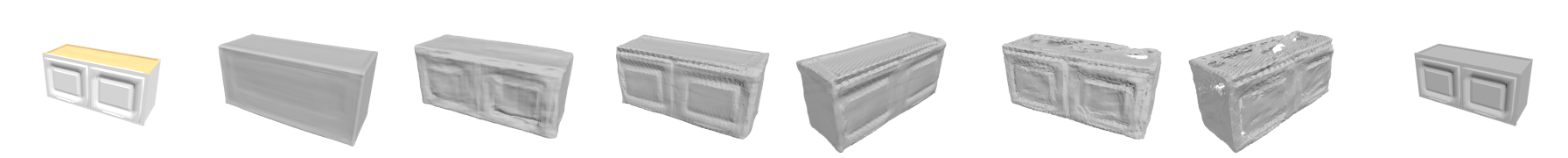}
\includegraphics[width=\textwidth]{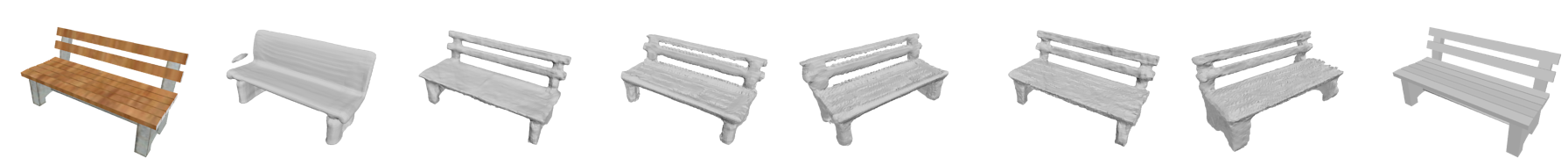}
\includegraphics[width=\textwidth]{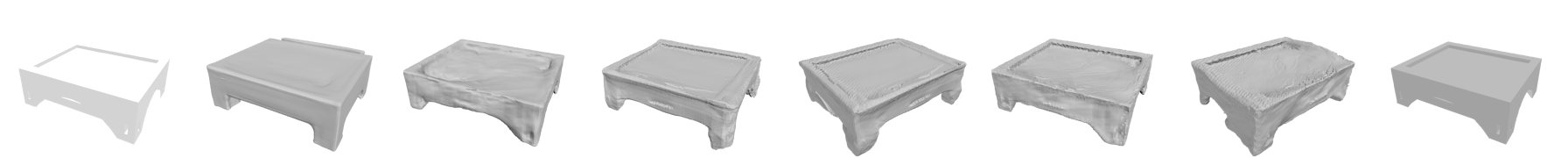}
\includegraphics[width=\textwidth]{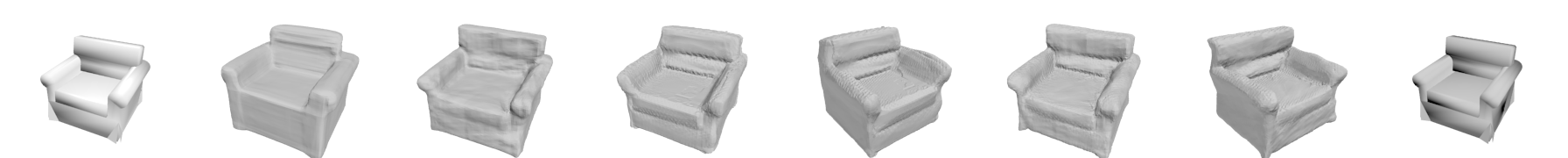}
\includegraphics[width=\textwidth]{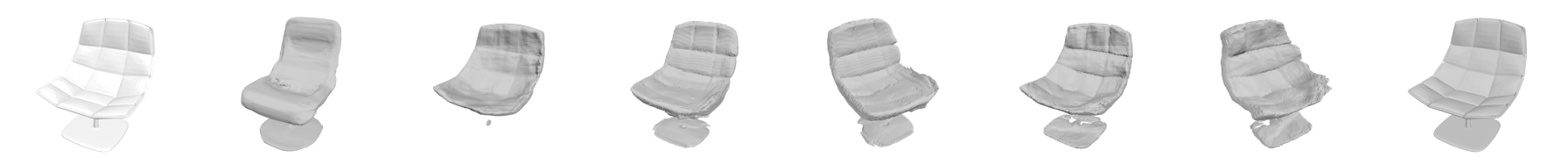}
\includegraphics[width=\textwidth]{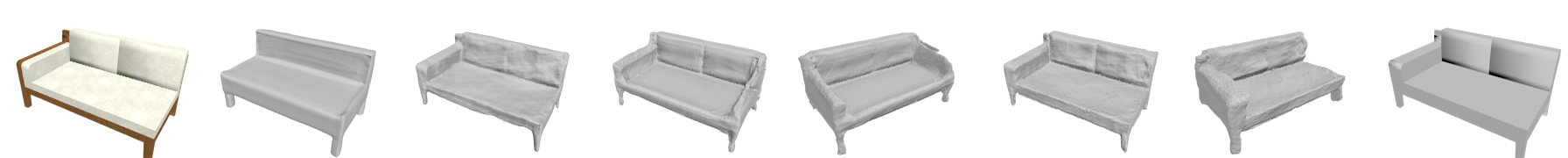}
\includegraphics[width=\textwidth]{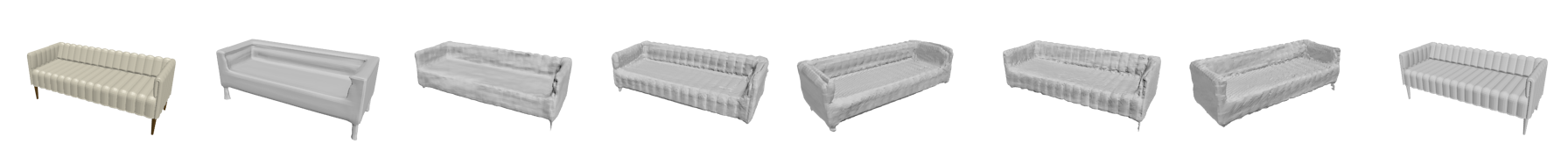}
\includegraphics[width=\textwidth]{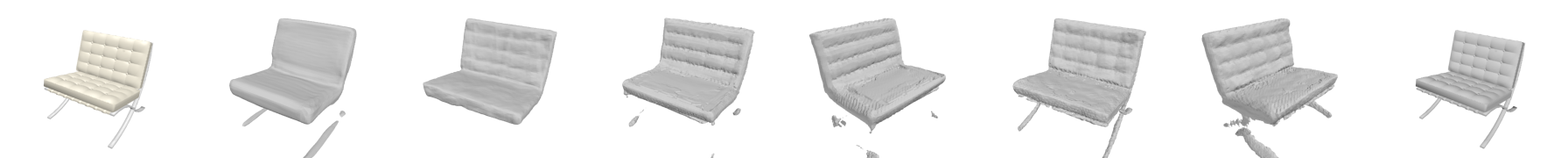}
\includegraphics[width=\textwidth]{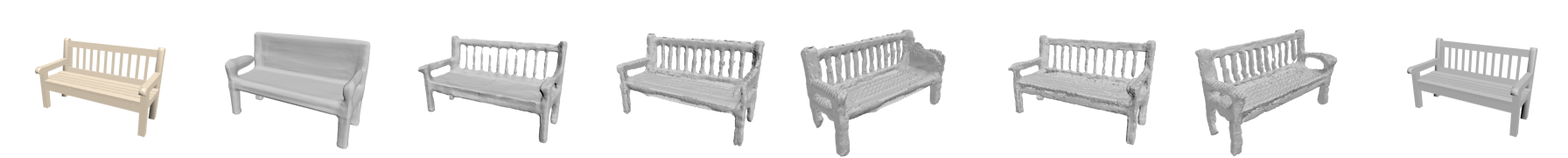}
\includegraphics[width=\textwidth]{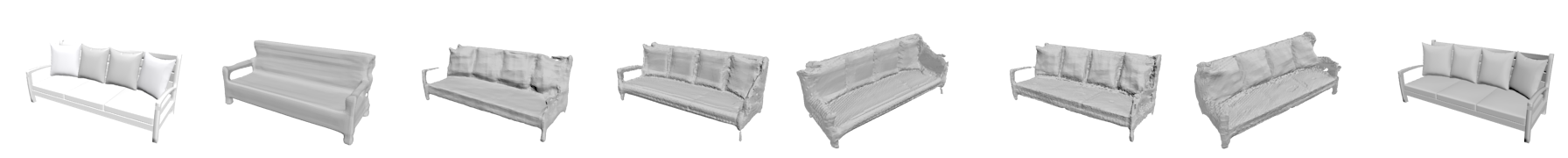}
\includegraphics[width=\textwidth]{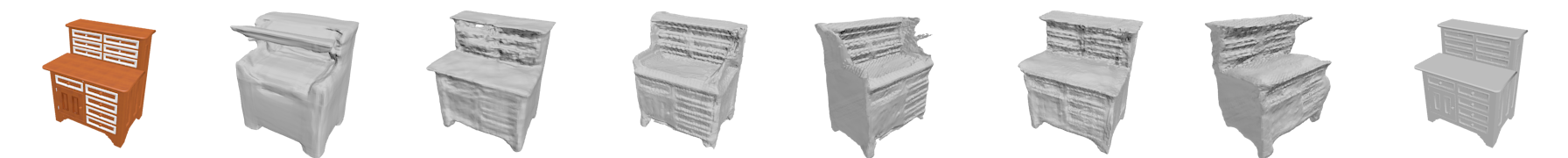}
\includegraphics[width=\textwidth]{images/supp_images/text.png}
\caption{More qualitative results. Two views of \ddi{} and \ddg{} are presented to show the reconstruction and recovered details.}
\label{comparison3}
\end{figure*}

\begin{figure*}
\centering
\includegraphics[width=\textwidth]{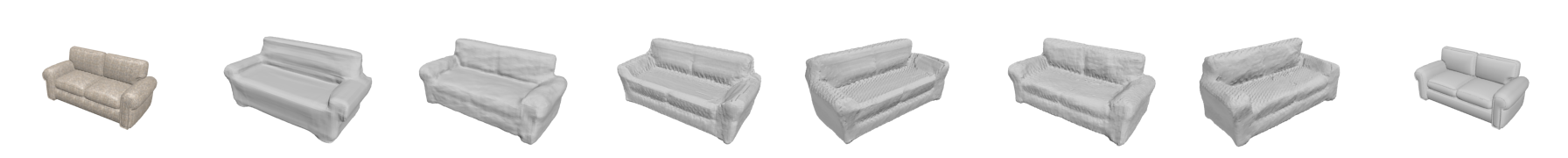}
\includegraphics[width=\textwidth]{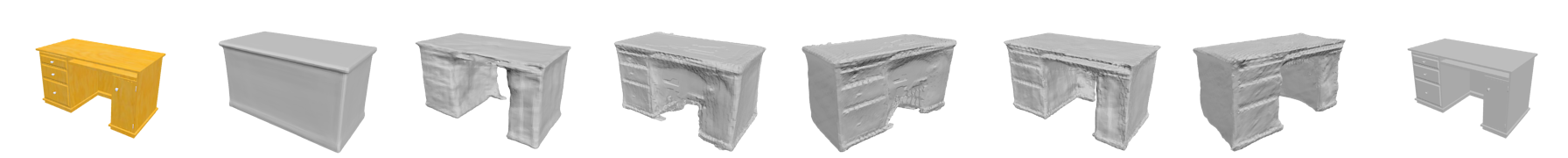}
\includegraphics[width=\textwidth]{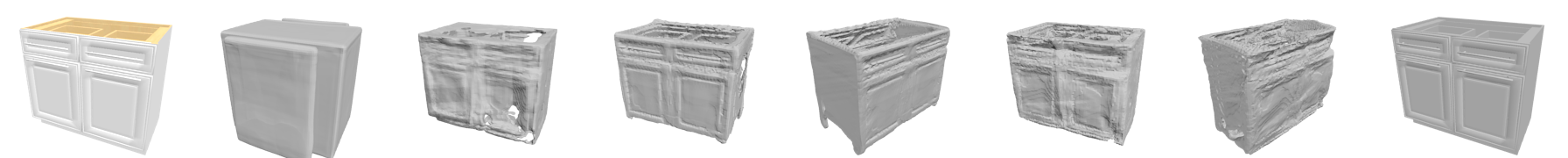}
\includegraphics[width=\textwidth]{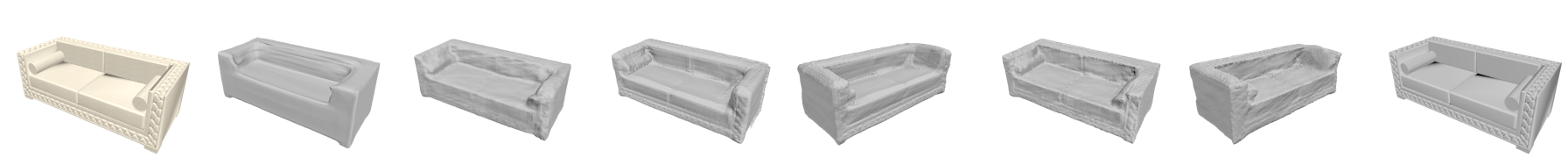}
\includegraphics[width=\textwidth]{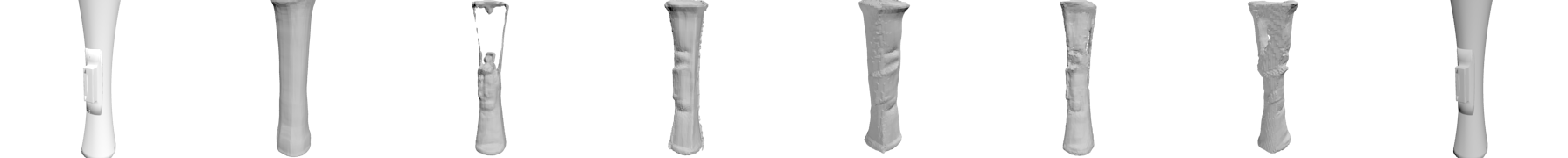}
\includegraphics[width=\textwidth]{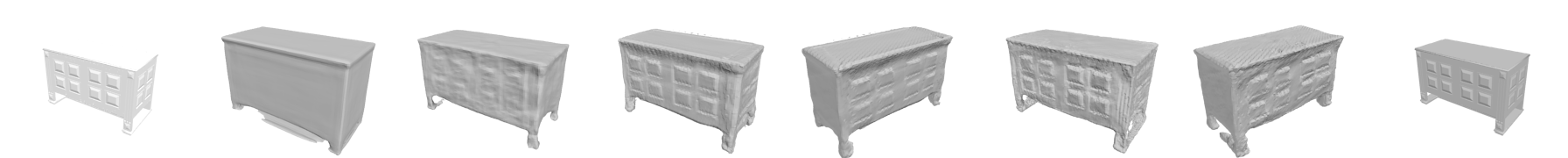}
\includegraphics[width=\textwidth]{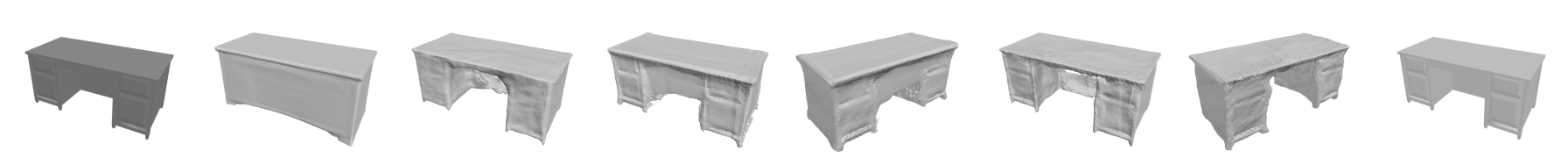}
\includegraphics[width=\textwidth]{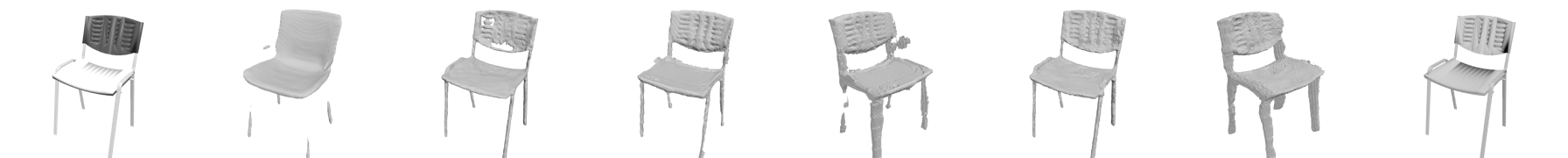}
\includegraphics[width=\textwidth]{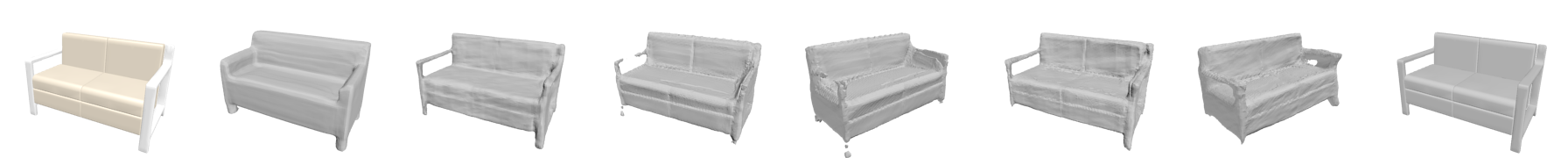}
\includegraphics[width=\textwidth]{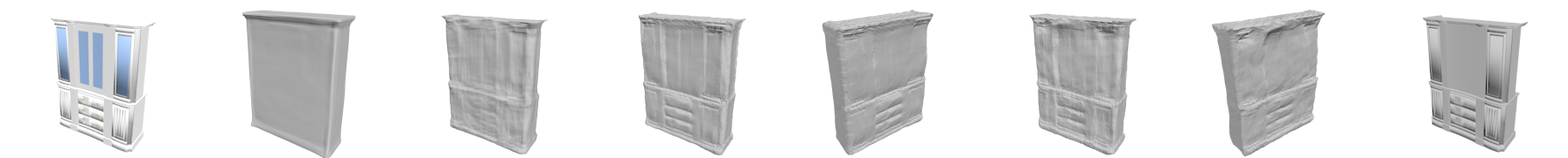}
\includegraphics[width=\textwidth]{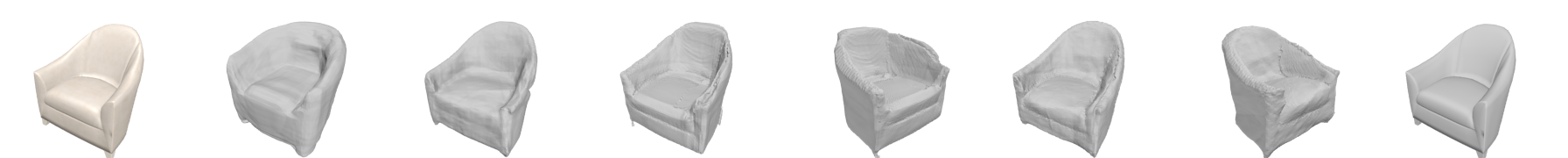}
\includegraphics[width=\textwidth]{images/supp_images/text.png}
\caption{More qualitative results. Two views of \ddi{} and \ddg{} are presented to show the reconstruction and recovered details.}
\label{comparison4}
\end{figure*}

\begin{figure*}
\centering
\includegraphics[width=\textwidth]{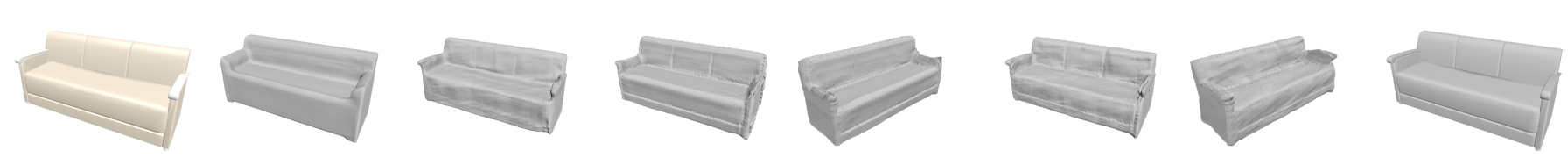}
\includegraphics[width=\textwidth]{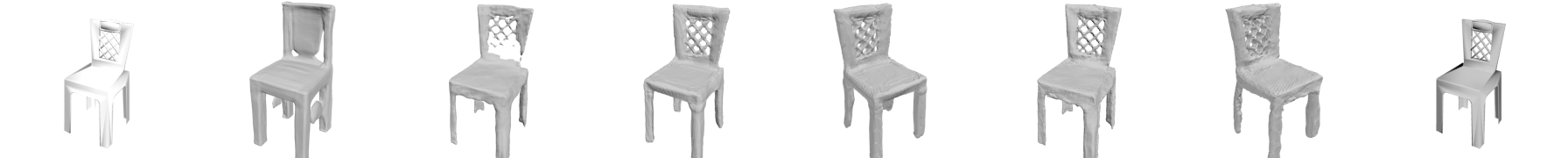}
\includegraphics[width=\textwidth]{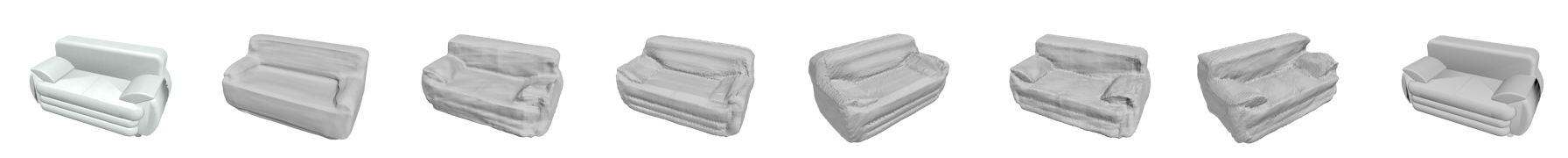}
\includegraphics[width=\textwidth]{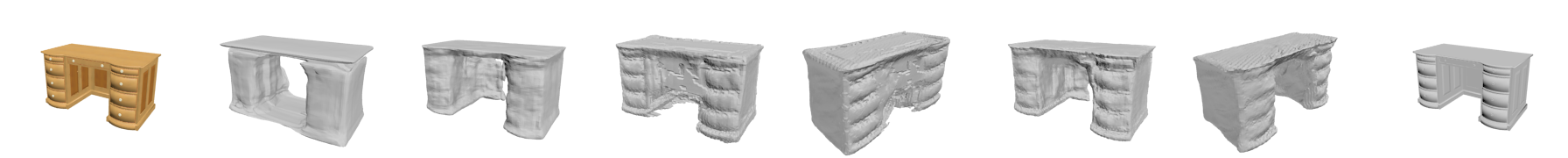}
\includegraphics[width=\textwidth]{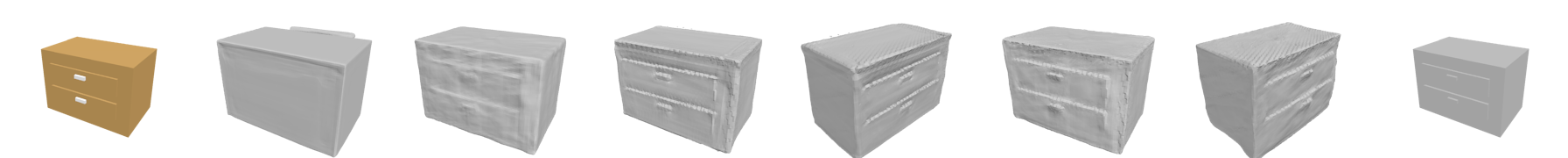}
\includegraphics[width=\textwidth]{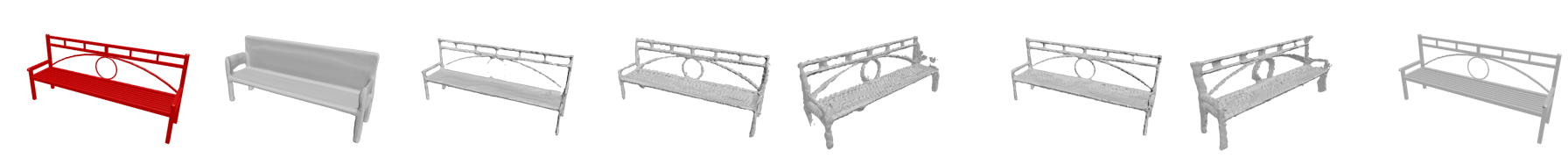}
\includegraphics[width=\textwidth]{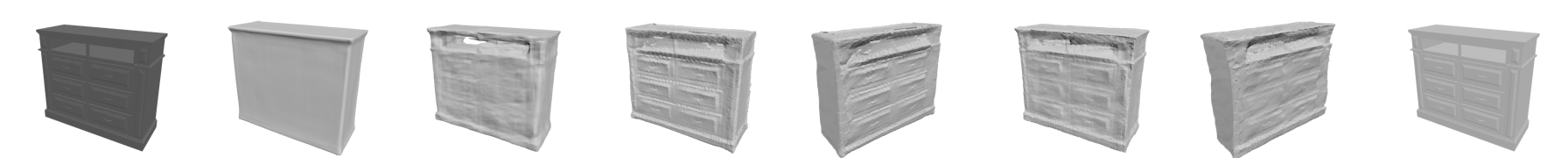}
\includegraphics[width=\textwidth]{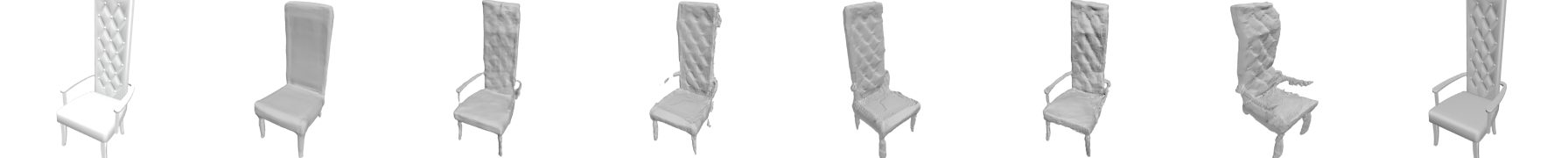}
\includegraphics[width=\textwidth]{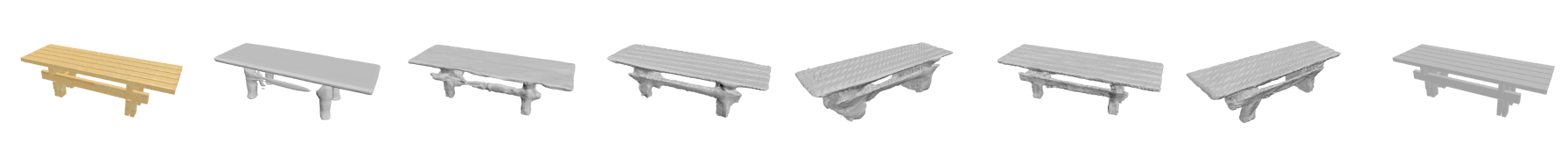}
\includegraphics[width=\textwidth]{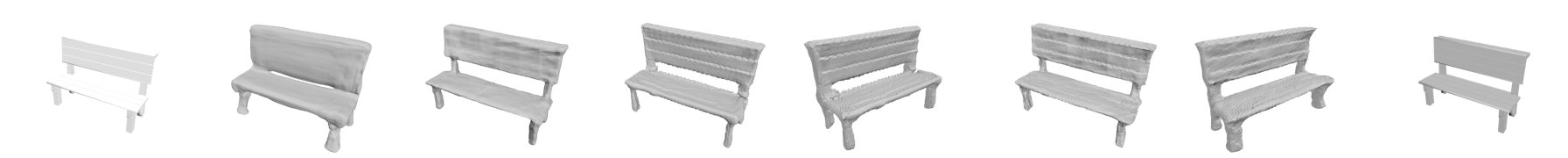}
\includegraphics[width=\textwidth]{images/supp_images/text.png}
\caption{More qualitative results. Two views of \ddi{} and \ddg{} are presented to show the reconstruction and recovered details.}
\label{comparison5}
\end{figure*}

Figure \ref{comparison1}, \ref{comparison2}, \ref{comparison3}, \ref{comparison4}, \ref{comparison5} show more qualitative results of single-view reconstruction. We mainly show the reconstructions of categories with clear details, such as chairs, sofas, cabinets, speakers. The results of \ddi{} recover the details while preserving the flatness of the other regions, which is preferred in the reconstruction scenarios.

{\small
\bibliographystyle{ieee_fullname}
\bibliography{references}
}